%% file: Hui_Okten.tex
\documentclass[review,onefignum,onetabnum]{siamart171218}

\input{ex_shared}

\ifpdf
\hypersetup{
  pdftitle={Derivative-based Shapley value for global sensitivity analysis and machine learning explainability},
  pdfauthor={Hui Duan, Giray \"{O}kten}
}
\fi

\myexternaldocument{ex_supplement}

\begin{document}
\nolinenumbers
\maketitle

\begin{abstract}
We introduce a new Shapley value approach for global sensitivity analysis and machine learning explainability. The method is based on the first-order partial derivatives of the underlying function. The computational complexity of the method is linear in dimension (number of features), as opposed to the exponential complexity of other Shapley value approaches in the literature. Examples from global sensitivity analysis and machine learning are used to compare the method numerically with activity scores, SHAP, and KernelSHAP.

\end{abstract}



\section{Introduction}
Global sensitivity analysis is the study of how uncertainty in the output of a model can be allocated to uncertainties in the model input. The applications of sensitivity analysis cover a wide range of disciplines, from natural sciences to engineering, and social sciences to mathematical sciences. Several methods for global sensitivity analysis have been introduced in the literature, including Sobol' sensitivity indices, derivative-based global sensitivity measures (DGSM), activity scores, and Shapley values. In this paper, we introduce a derivative-based Shapley value (DerSHAP) approach for global sensitivity analysis and machine learning explainability.

The paper is organized as follows. In Section \ref{sec:gsa} we discuss Sobol' sensitivity indices, derivative-based global sensitivity measures, and activity scores. In Section \ref{sec:shapley} we review the Shapley value, and in Section \ref{sec:dershap} we introduce our method, the derivative-based Shapley value (DerSHAP). In Section \ref{sec:results} we use DerSHAP to perform global sensitivity analysis of an Ebola model and an HIV model, and compare it numerically with the activity score approach. We also apply DerSHAP to two datasets from machine learning, Amazon stock data and Boston housing data, and compare it with SHAP and KernelSHAP. Conclusions follow in the last section.

\section{Global Sensitivity Analysis}
\label{sec:gsa}
In this section, we review some of the background material from global sensitivity analysis. We will give a review of Sobol' sensitivity indices, derivative based global sensitivity measures (DGSM), and activity scores. 
\subsection{Sobol' Sensitivity Indices}
Consider a $d$ dimensional input vector $\textbf{x} = (x_1, \ldots, x_d)$ with the index set $D = \{1,2,\ldots, d\}$.  Let  $f(\textbf{x})$ be a square-integrable function defined on $(0,1)^d$. The ANOVA decomposition of $f(\textbf{x})$ is expressed as
$$
f(\textbf{x}) = \sum_{u \subseteq D}f_u(\textbf{x}^u),
$$
where $f_u(\textbf{x}^u)$ is the component function that only depends on $\textbf{x}^u$. For the empty set, we have
$f_{\emptyset} = \int f(\textbf{x}) d\textbf{x}$. 

If we assume $\textbf{x}$ has uniform distribution on $(0,1)^d$, we can write
$$
\mathbb{E}[f(\textbf{x})] = \int_{(0,1)^d} f(\textbf{x}) d\textbf{x},
$$
and
$$
\text{Var}(f(\textbf{x}))=\sigma^2 = \int_{(0,1)^d} f^2(\textbf{x}) d\textbf{x} - \mathbb{E}[f(\textbf{x})]^2.
$$
Due to the orthogonality of the ANOVA decomposition, the variance of $f$ can be written as  
$$
\sigma^2 =\sum_{u \subseteq D}\sigma_u^2,
$$
where $\sigma_u^2$ is the variance for the component function $f_u$:
 $$
 \sigma_u^2 = \int_{(0,1)^d} f_u^2(\textbf{x})d\textbf{x} - \left(\int_{(0,1)^d} f_u(\textbf{x}) d\textbf{x}\right)^2 = \int_{(0,1)^d} f_u^2(\textbf{x})
d\textbf{x}.
 $$
The Sobol' sensitivity indices for the subset $u$ are defined from the variances of component functions
$$
\underline{S}_u = \frac{1}{\sigma^2}\sum_{v \subseteq u} \sigma_v^2 = \frac{\underline{\tau}_u}{\sigma^2}
\text{       and       }
\overline{S}_u = \frac{1}{\sigma^2}\sum_{v \bigcap u \neq \emptyset} \sigma_v^2 = \frac{\overline{\tau}_u}{\sigma^2}, 
$$ 
where $\underline{S}_u $ is called the lower Sobol' sensitivity index (or, the main effect) and $\overline{S}_u$ is called the upper Sobol' sensitivity index (or, the total effect).  If $\underline{S}_u$ is close to 1, then the parameters $\textbf{x}^u$ are viewed as very important to the model. In practical applications, we often consider the case when $u$ is a singleton $\{i\}$. If $\overline{S}_{\{i\}}$ is close to 0, we consider $x_i$ to be an unimportant variable, and freeze it to its mean value to simplify the model $f$.

Sobol' sensitivity indices can be generalized to independent random vectors $\textbf{x} = (x_1,\ldots, x_d)$ with any distribution functions $F_1(x_1), \ldots, F_d(x_d)$. See Kucherenko et. al (2012) \cite{kucherenko2012estimation} for details.

\subsection{Derivative Based Global Sensitivity Measures}

If the partial derivatives of $f$ exist, computationally more efficient global sensitivity measures can be obtained.  For example, Campolongo et al. (2007) \cite{campolongo2007effective} introduced a global sensitivity measure based on $\int_{(0,1)^d}
\left | \frac{ \partial f(\textbf{x})}{\partial x_i}\right | d\textbf{x}$, and Sobol' \& Kucherenko (2009) \cite{dgsmSobol} introduced a global sensitivity measure based on $\int_{(0,1)^d} (\frac{\partial f(\textbf{x})}{\partial x_i})^2 d\textbf{x}$. Here we will discuss the latter measure further.
The derivative based global sensitivity measure (DGSM) of Sobol' \& Kucherenko (2009) \cite{dgsmSobol} is defined as
\begin{equation}
    v_i = \int_{(0,1)^d}(\frac{\partial f(\textbf{x})}{\partial x_i})^2 d\textbf{x} = \mathbb{E}[(\frac{\partial f(\textbf{x})}{\partial x_i})^2].
    \label{dgsm1}
\end{equation}
DGSM can be estimated by Monte Carlo as
\[
v_i \approx \hat{v}_i = \frac{1}{N}\sum_{j=1}^N (\frac{\partial f(\textbf{x}^{(j)})}{\partial x_i})^2,
\]
where $\textbf{x}^{(j)},j=1,\ldots,N$ is a random sample from the uniform distribution on $(0,1)^d$.

Sobol' (2011) \cite{sobol2011derivative} showed that if $f(\x)$ is a linear function on each of its components $x_i$, then
\begin{equation}
\overline{S}_i = \frac{1}{12}\frac{v_i}{\sigma^2}. 
\label{relationship1}
\end{equation}
In general, the following inequality was obtained in Sobol' \& Kucherenko (2009) \cite{dgsmSobol}
\begin{equation}
\overline{S}_i \leq \frac{1}{\pi^2} \frac{v_i}{\sigma^2}. 
\label{ineq_upper}
\end{equation}

DGSM can be generalized to nonuniform distributions, where the random vector $\x=(x_1,\ldots x_d)$ has independent components with marginal distributions $F_1,\ldots,F_d$, and each $F_i$ has a corresponding density function $f_i$. Sobol' \& Kucherenko (2009) \cite{dgsmSobol} and Kucherenko and Iooss (2014) \cite{kucherenko2014derivative} obtained several results in this generalized setting. In particular, the following generalization of inequality  (\ref{ineq_upper}) was obtained in \cite{kucherenko2014derivative}:
\begin{equation}
\overline{S}_i \leq D(F_i) \frac {v_i}{\sigma^2},
\label{eq_dgsm}
\end{equation}
where
$$
D(F_i) = 4\left[\sup_{x\in \mathbb{R}} \frac{\min(F_i(x), 1-F_i(x))}{f_i(x)}\right]^2.
$$

\subsection{Activity Scores}
The activity score was introduced by Constantine and Diaz (2017) \cite{constantine2017global}, and it is based on the active subspace method (Constantine et. al  (2014) \cite{constantine2014active}). The active subspace method finds the important directions of a function using the gradient information, and uses these important directions to reduce the domain of the function to a subspace.  The activity score is a global sensitivity measure obtained from this information. We follow \cite{constantine2017global} to describe the method. 

Consider a square-integrable function $f(\x)$ defined on $(-1,1)^d$, with finite partial derivatives. Assume the partial derivatives of $f$ are square-integrable. Let $p(\x)$ be the uniform probability density function on $(-1,1)^d$. The gradient of $f$, $\nabla f(\x) \in \mathbb{R}^d$, is given by
$$\nabla f(\x) = \left[\frac{\partial f(\textbf{x})}{\partial x_1},\frac{\partial f(\textbf{x})}{\partial x_2},\ldots, \frac{\partial f(\textbf{x})}{\partial x_d}\right]^T. 
$$
Define the matrix $C$ as
\begin{equation}
C =  \mathbb{E}[\nabla f(\textbf{x})\nabla f(\textbf{x})^T],
\label{eigen_eqn}
\end{equation}
where the expectation is computed using the density function $p$. Consider the eigenvalue decomposition of $C$
\[
C=\textbf{W}\Lambda \textbf{W}^T,
\]
where $\textbf{W} = [\textbf{w}_1,\ldots, \textbf{w}_d]$ is the $d\times d$ orthogonal matrix of eigenvectors, and $\Lambda = \text{diag}(\lambda_1,\ldots, \lambda_d)$ with $\lambda_1 \geq \ldots \geq \lambda_d \geq 0$ is the diagonal matrix of eigenvalues in descending order. Matrix $C$ can be approximated using the Monte Carlo method:
$$
C \approx \hat{C} = \frac{1}{N}\sum_{j=1}^N(\nabla_{\x} f(\x^{({j})}))(\nabla_{\x} (f(\x^{(j)}))^T,
$$
where the $\x^{(1)},\ldots,\x^{(N)}$ is a random sample of size $N$ from the density $p$.

If the eigenvalues $\lambda_{m+1},\ldots,\lambda_d$ are sufficiently small, then the active subspace method approximates $f(\textbf{x})$ with a lower dimensional function $g(\textbf{W}_1)$ where $\textbf{W}_1$ is the $d\times m$ matrix containing the first $m$ eigenvectors. The dimension of $g$ is $m$, whereas the dimension of $f$ is $d$.

The activity score for the $i$th parameter is defined as
\begin{equation}
\alpha_i(m) = \sum_{j=1}^{m} \lambda_j w_{ij}^2,
\label{ac_1}
\end{equation}
where $\textbf{w}_j=[w_{1j}, \ldots, w_{dj}]^T$ is the $j$th eigenvector, $m\leq d$, and $i = 1,\ldots, d$. 

Constantine and Diaz (2017) \cite{constantine2017global}  showed that the activity scores are bounded by DGSM 
$$
\alpha_i (m) \leq v_i ,\text{       } i=1,\ldots,d,
$$
and the inequality becomes an equality if $m = d$.

The following inequality between the Sobol' sensitivity index and activity scores was also established in  \cite{constantine2017global}:
$$
\overline{S}_i\leq \frac{1}{4\pi^2}\frac{\alpha_i(m) + \lambda_{m+1}}{\sigma^2}.
$$

This result can be generalized to non-uniform distributions over $R^d$ using Eqn. (\ref{eq_dgsm}). 
\begin{lemma}
For any random vector $\x$ of independent components with marginal distributions $F_i$ and density $f_i$, $i=1,\ldots,d$, distributions $F_1, \ldots, F_d$, we have 
$$
\overline{S}_i\leq D(F_i)\frac{(\alpha_i(m)+\lambda_{m+1})}{\sigma^2}.
$$ 
\end{lemma}

\section{Shapley Values}
\label{sec:shapley}
Shapley (2016) \cite{shapley201617} introduced the Shapley value to solve the question of distributing a reward among multiple team members fairly in cooperative game theory. Owen (2014) \cite{owen2014sobol} used the Shapley value approach to quantify the importance of input variables of a function to the function output. This approach is particularly valuable when the input variables are dependent (Owen \& Prieur (2017) \cite{owen2017shapley}). 

Consider a function $f$ of $d$ variables, $f(x_1,\ldots,x_d)$, and let $\mathcal{D}=\{1,2,\ldots, d\}$ be the index set. Let $\text{imp}$ be a function from the powerset of $\mathcal{D}$ to positive real numbers, with the interpretation that for any subset $u$ of $\mathcal{D}$, $\text{imp}(u)$ is a measure of importance of input variables with indices in $u$, and $\text{imp}(\emptyset)=0$. The question we want to answer is how much of the importance of the whole set of input variables can be attributed to each input variable. To this end, let $\phi_i$ denote the contribution of input variable $x_i$ to $\text{imp}(\mathcal{D})$ (in the terminology of cooperative game theory, this function is called the Shapley value). Shapley showed that for any given function $\text{imp}$, there is a corresponding function $\phi_i$ satisfying the following conditions:

\begin{enumerate}
\item Efficiency: $\sum_{i =1}^d\phi_i = \text{imp}(\mathcal{D})$;
\item Dummy: If $\text{imp}(u\cup \{i\}) = \text{imp}(u)$ for all $u\subseteq \mathcal{D}$, then $\phi_i = 0$;
\item Symmetry: If $\text{imp}(u\cup \{i\}) = \text{imp}(u\cup \{j\})$ for all $u\subseteq \mathcal{D}$ such that $u\cap \{i,j\} = \emptyset$, then $\phi_i = \phi_j$; 
\item Additivity: If $\text{imp},\text{imp}' $ are two measures of importance with corresponding functions $\phi$, $\phi'$, then the importance measure $\text{imp} + \text{imp}'$ has the corresponding function $\phi_i + \phi_i'$.
\end{enumerate}

This unique solution is given by
\begin{equation}
\phi_i = \frac{1}{d}\sum_{u\subseteq -\{i\}} {d-1 \choose |u|}^{-1} (\text{imp}(u\cup \{i\}) - \text{imp}(u)),
\label{e0}
\end{equation}
where $|u|$ denotes the cardinality of $u$, and $-\{i\}$ is the complement of $\{i\}$.

For black-box models, Lundberg and Lee (2017) \cite{lundberg2017unified} proposed the Shapley additive explanation (SHAP). This method uses conditional expectations to represent the imp function and has gained popularity in machine learning (Broeck et al. (2022) \cite{van2022tractability}), despite some authors having raised questions about its explainability (Kumar et al. (2020) \cite{kumar2020problems}). Lundberg et al. (2018) \cite{lundberg2018consistent} introduced TreeSHAP to compute the Shapley values for tree-based machine learning models with reduced complexity. Mase et al. (2019) \cite{mase2019explaining} proposed the cohort Shapley to measure the similarity of a specific subject to the target subject. Seiler et al. (2021) \cite{seiler2021makes} defined uniqueness Shapley based on the size of the cohort. They aimed to interpret the importance of a variable in identifying a specific subject from a set of data in the machine learning models.


\section{Shapley Value Based on Derivatives}
\label{sec:dershap}
 In this section we introduce an importance function, $\text{imp}(u)$, for subsets $u$ of $\mathcal{D}$ based on the partial derivatives of $f$, similar in spirit, to the DGSM and activity score measures of global sensitivity analysis. We assume that the first-order partial derivatives of $f$ are finite, and square-integrable. To simplify the notation, we denote the partial derivative of $f(\textbf{x})$ with respect to $x_i$ as
 $$
 \frac{\partial f(\textbf{x})}{\partial x_i}=\partial f_i. 
 $$
Define the \textit{derivative-based importance} of subset $u$ as follows:
\begin{equation}
\text{imp}(u) = \sum_{i\in u}\sum_{j\in u,j\geq i}\left|\mathbb{E}[\partial f_i\partial f_j]\right| = \sum_{i\in u}\mathbb{E}[\partial f_i^2 ]+ \sum_{i,j \in u, j>i}\left|\mathbb{E}[\partial f_i\partial f_j]\right|.
\end{equation}
Note that the first summation of the right-hand side can be written in terms of the activity scores
\[
\sum_{i\in u} \mathbb{E}[\partial f_i^2 ]=\sum_{i\in u}\alpha_i(d),
\]
and thus we can write
$$
\text{imp}(u)  = \sum_{i\in u}\alpha_i(d)+ \sum_{i,j \in u, j>i}\left |\mathbb{E}[\partial f_i\partial f_j]\right |.
$$
For the empty set, we set $\text{imp}(\emptyset) = 0$, and for the full index set, $\text{imp}(\mathcal{D})$ is the sum of entries of the upper triangular matrix of $|C|$ (see Eqn. (\ref{eigen_eqn})):
$$
\text{imp}(\mathcal{D}) = \sum_{i=1}^d\sum_{j\geq i}^d\left | \mathbb{E}[\partial f_i\partial f_j]\right|. 
$$

\begin{theorem}
The Shapley value for the derivative-based importance function is
\begin{equation*}
\phi_i 
= \mathbb{E}[\partial f_i^2]+ \frac{1}{2} \sum_{j=1,j\neq i}^d\left | \mathbb{E}[\partial f_i\partial f_j]\right|.
\label{eq:derS}
\end{equation*}
\end{theorem}
\begin{proof}
The efficiency condition follows from
\begin{align*}
\sum_{i=1}^d \phi_i 
=\sum_{i=1}^d\mathbb{E}[\partial f_i^2] + \sum_{i=1}^{d-1}\sum_{j=i+1}^d \left |\mathbb{E}[\partial f_i\partial f_j]\right | 
=\text{imp}(\mathcal{D}).
\end{align*}
To prove the dummy condition, assume $\text{imp}(u \cup \{i\}) =\text{imp}(u)$ for all subsets $u$. Also assume we can write $\mathcal{D}= v \cup w \cup \{i\}$ where $v,w$ are nonempty disjoint sets not containing $i$. Observe that $\text{imp}(v \cup \{i\}) =\text{imp}(v)$ implies
\[
\mathbb{E}[\partial f_i^2] + \sum_{k<i,k\in v}\left | \mathbb{E}[\partial f_k \partial f_i] \right|+ \sum_{k>i,k\in v} \left | \mathbb{E}[\partial f_i \partial f_k] \right|
= \mathbb{E}[\partial f_i^2] + \sum_{k\in v} \left | \mathbb{E}[\partial f_k \partial f_i ]\right | =0.
\]
Likewise, $\text{imp}(w \cup \{i\}) =\text{imp}(w)$ implies
\[
\mathbb{E}[\partial f_i^2] + \sum_{k\in w} \left | \mathbb{E}[\partial f_k \partial f_i] \right |=0.
\]
Adding the last two equations we obtain
\[
2 \phi_i = 2 \mathbb{E}[\partial f_i^2]  + \sum_{k \neq i, k\in \mathcal{D}}\left | \mathbb{E}[ \partial f_k \partial f_i ]\right |=0.
\]

To prove symmetry assume $\text{imp}(u\cup\{k_1\}) =  \text{imp}(u\cup\{k_2\})$ for all subsets $u$ that do not contain $k_1$ and $k_2$. Assume we can write $\mathcal{D}= v \cup w \cup \{k_1,k_2\}$ where $v,w$ are nonempty disjoint sets not containing $k_1$ and $k_2$. (If $v$ and $w$ are the empty set and a singleton, the condition holds trivially.) Note that $\text{imp}(v\cup\{k_1\}) =  \text{imp}(v\cup\{k_2\})$ implies
\[
\mathbb{E}[\partial f_{k_1}^2]+\sum_{i\in v}\left | \mathbb{E}[\partial f_{k_1}\partial f_i]\right |=
\mathbb{E}[\partial f_{k_2}^2]+\sum_{i\in v}\left | \mathbb{E}[\partial f_{k_2}\partial f_i]\right |
\]
and $\text{imp}(w\cup\{k_1\}) =  \text{imp}(w\cup\{k_2\})$ implies
\[
\mathbb{E}[\partial f_{k_1}^2]+\sum_{i\in w}\left | \mathbb{E}[\partial f_{k_1}\partial f_i]\right |=
\mathbb{E}[\partial f_{k_2}^2]+\sum_{i\in w}\left | \mathbb{E}[\partial f_{k_2}\partial f_i]\right | .
\]
Adding these equations we get $\phi_{k_1}= \phi_{k_2}$.
\end{proof}

The derivative-based Shapley (DerSHAP) value involves the summation of $d$ terms where each summand is an expectation that can be approximated using Monte Carlo. Therefore the complexity of the method is $O(dN)$ where $N$ is the Monte Carlo sample size. In contrast, the original Shapley value formulation requires a summation of $2^d$ terms (Eqn. (\ref{e0})). The Shapley values introduced by Owen \& Prieur (2017)\cite{owen2017shapley}, Song et al. (2016) \cite{song2016shapley}, and Lundberg and Lee (2017) \cite{lundberg2017unified} all have exponential complexity.
KernelSHAP (Lundberg and Lee (2017) \cite{lundberg2017unified}) provides a more efficient algorithm than SHAP, where it approximate SHAP values using linear regression. Nevertheless, it still has exponential complexity. TreeSHAP(Lundberg et al. (2018) \cite{lundberg2018consistent}) employs the same important function as SHAP but has polynomial complexity $O(TLD^2)$ for tree-based models, where $L$ is the number of leaves for an ensemble tree, $T$ is the number of trees in the ensemble, and $D$ is the maximum tree depth.



If the eigenvalues $\lambda_{k+1},\ldots,\lambda_d$ of the matrix $|C|$ (Eqn. (\ref{eigen_eqn})) are small, then we can approximate the Shapley values by setting the small eigenvalues to zero. Let the eigenvalue decomposition of $|C|$ be given as $|C|=\textbf{W}\Lambda \textbf{W}^T$, and write
$$
|C|= \sum_{i=1}^d \lambda_iw_iw_i^T,
$$
where $w_i$ are the orthonormal eigenvectors. Assume $\lambda_i < \epsilon$ for $i=k+1,\ldots,d$, and define 
$$
\tilde{C} = \sum_{i=1}^k \lambda_i w_iw_i^T, 
$$
where $k\leq d$. 
The diagonal entries of $|C| - \tilde{C}$ are
$$
[\left |C\right |-\tilde{C}]_{j,j} = \sum_{i=k+1}^d \lambda_iw_{j,i}^2 \leq \sum_{i=k+1}^d\lambda_i.
$$
The Frobenius norm of $|C| - \tilde{C}$ satisfies
$$
\|\left | C\right | - \tilde{C}\|_F^2 = \| \sum_{i=k+1}^d \lambda_i w_iw_i^T\|_F^2 = \sum_{i=k+1}^d \lambda_i^2 \leq (d-k) \epsilon^2. 
$$

Let $\Phi$ be the vector of derivative-based Shapley values, that is, $\Phi=[\phi_1,\ldots,\phi_d]^T$. Let $e$ be the column vector of 1's. We have
$$
\Phi = \frac{1}{2} |C| e + \frac{1}{2} \diag(|C|) e. 
$$
Define $\tilde{\Phi}$ as the vector of ``approximate" Shapley values, where the matrix $|C|$ is replaced by $\tilde{C}$:
$$
\tilde{\Phi} = \frac{1}{2} \tilde{C} e + \frac{1}{2} \diag(\tilde{C}) e.
$$
The error between $\Phi$ and $\tilde{\Phi}$ is 
$$
\Phi - \tilde{\Phi} = \frac{1}{2} (|C|-\tilde{C}) e + \frac{1}{2} \diag(|C|-\tilde{C}) e,
$$
and can be bounded by
\begin{align*}
\|\Phi - \tilde{\Phi}\|_2 &= \|\frac{1}{2} (|C|-\tilde{C})  e + \frac{1}{2}  \diag(|C|-\tilde{C}) e\|_2\\
&\leq \frac{1}{2}  \|(|C|-\tilde{C}) e\|_2 + \frac{1}{2}  \|\diag(|C|-\tilde{C}) e\|_2\\
&\leq \frac{1}{2}\sqrt{d-k}  \epsilon   \sqrt{d} + \frac{1}{2}(d-k) \epsilon  \sqrt{d}\\
&\leq (d-k) \epsilon \sqrt{d},
\end{align*}
where $k\leq d$, and in the second inequality we used the fact $\|(|C|-\tilde{C}) e\|_2 \leq \| \left |C\right |-\tilde{C}\|_F  \|e\|_2$.


\section{Numerical Results}
\label{sec:results}
In this section we apply the derivative-based Shapley value (DerSHAP) to perform global sensitivity analysis of an Ebola model and an HIV model. We will compare the method with the activity scores approach in these applications. In both applications the model parameters have independent random inputs, and thus the activity scores approach is applicable. We will then apply DerSHAP to two datasets from machine learning, Amazon stock data and Boston housing data, and compare DerSHAP with SHAP and KernelSHAP. In these problems the input parameters are not independent. In all the numerical results, we report the normalized activity scores and normalized Shapley values. Normalization simply means dividing each value (activity score or Shapley value) by the sum of all values. The sum of the normalized activity scores and the sum of the normalized Shapley values are equal to one. 

SHAP assumes feature independence, and KernelSHAP assumes feature independence and model linearity. In the numerical examples, we use SHAP Python package (2021)\footnote{\url{https://github.com/slundberg/shap}} to implement both SHAP and KernelSHAP.

\subsection{The Ebola Model}
This is a modified SEIR model for the spread of Ebola in Western Africa. A detailed description of the model is given in Diaz et al. (2018) \cite{diaz2018modified} where the activity score method was used to perform the global sensitivity analysis of the model. 

The SEIR model is a system of seven differential equations which describe the dynamics of disease within a population. We are interested in estimating the basic reproduction number, which is a metric measuring how many new cases of disease each current case causes. Using the steady state solution and the next generation matrix, Diaz et al. (2018) \cite{diaz2018modified} shows that the basic reproduction number $R_0$ is
$$
R_0 = \left(\beta_1 + \frac{\beta_2\rho_1\gamma_1}{\omega} + \frac{\beta_3}{\gamma_2}\psi\right)/\left(\gamma_1 + \psi \right).
$$

\begin{figure}[h]
\centering
\begin{subfigure}[b]{.45\textwidth}
    \centering
    \includegraphics[width=1\textwidth]{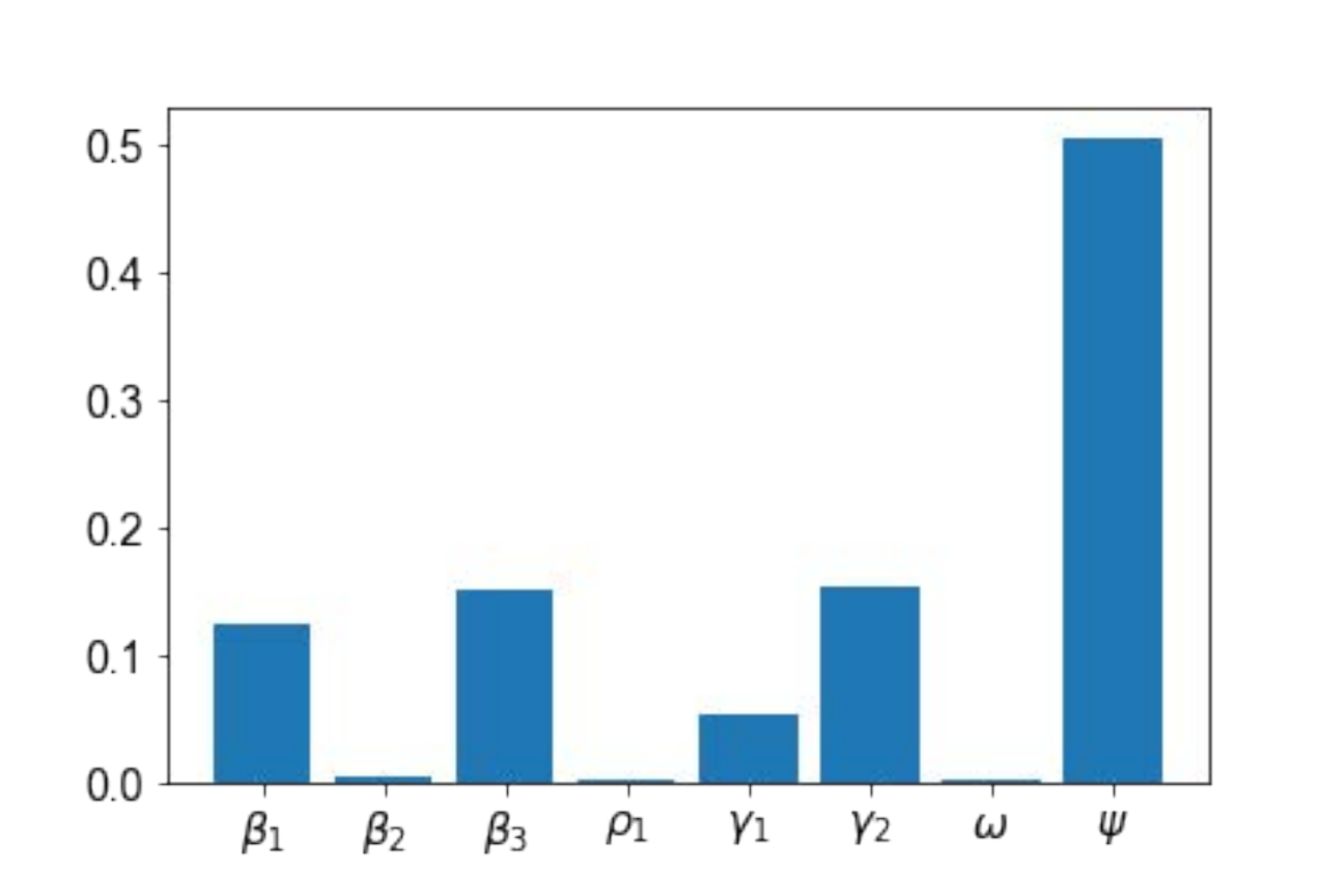}
    \subcaption{Activity scores - Liberia}
\end{subfigure}%
\begin{subfigure}[b]{.45\textwidth}
    \centering
    \includegraphics[width=1\textwidth]{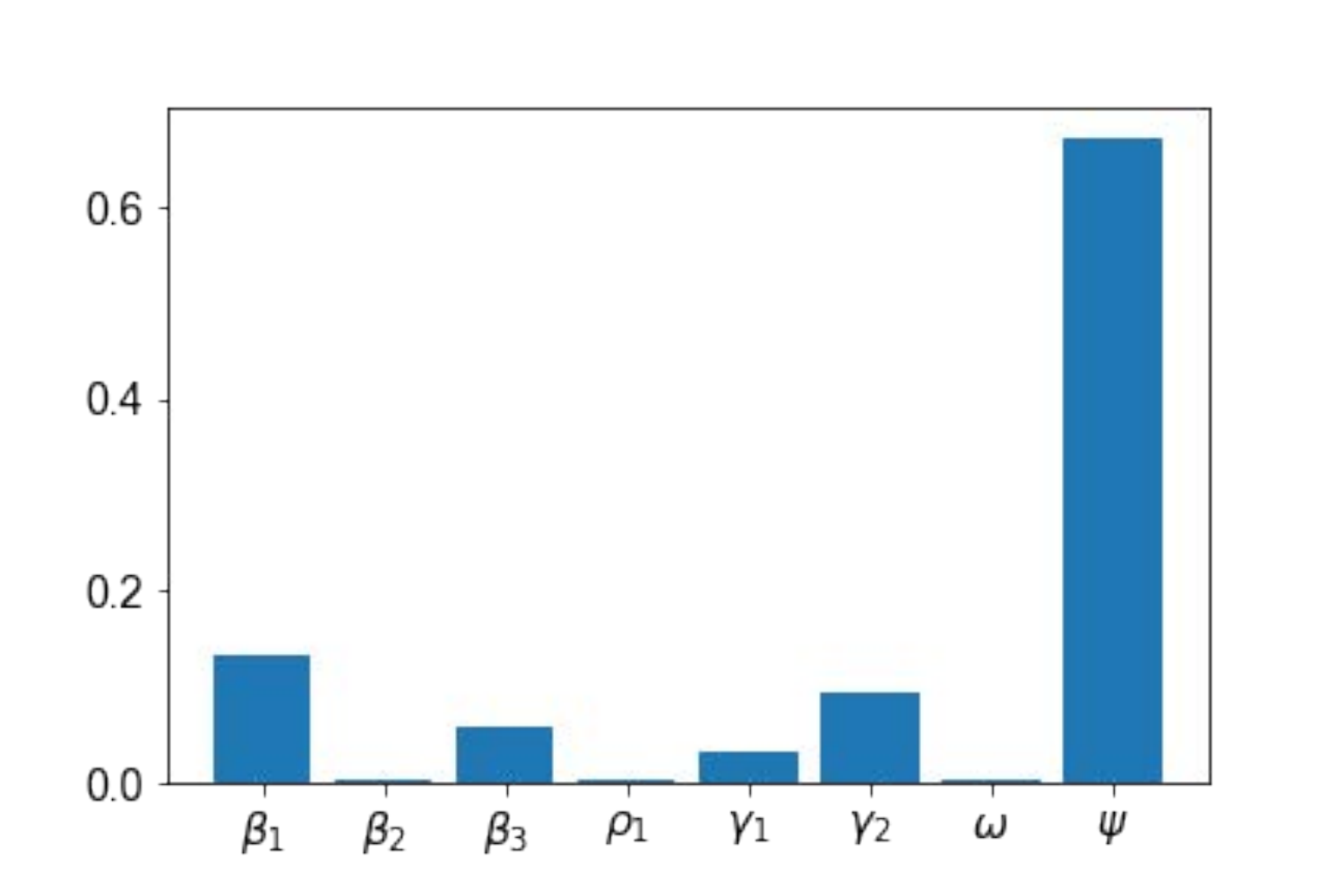}
    \subcaption{Activity scores - Sierra Leone}
\end{subfigure}
\begin{subfigure}[b]{.45\textwidth}
    \centering
    \includegraphics[width=1\textwidth]{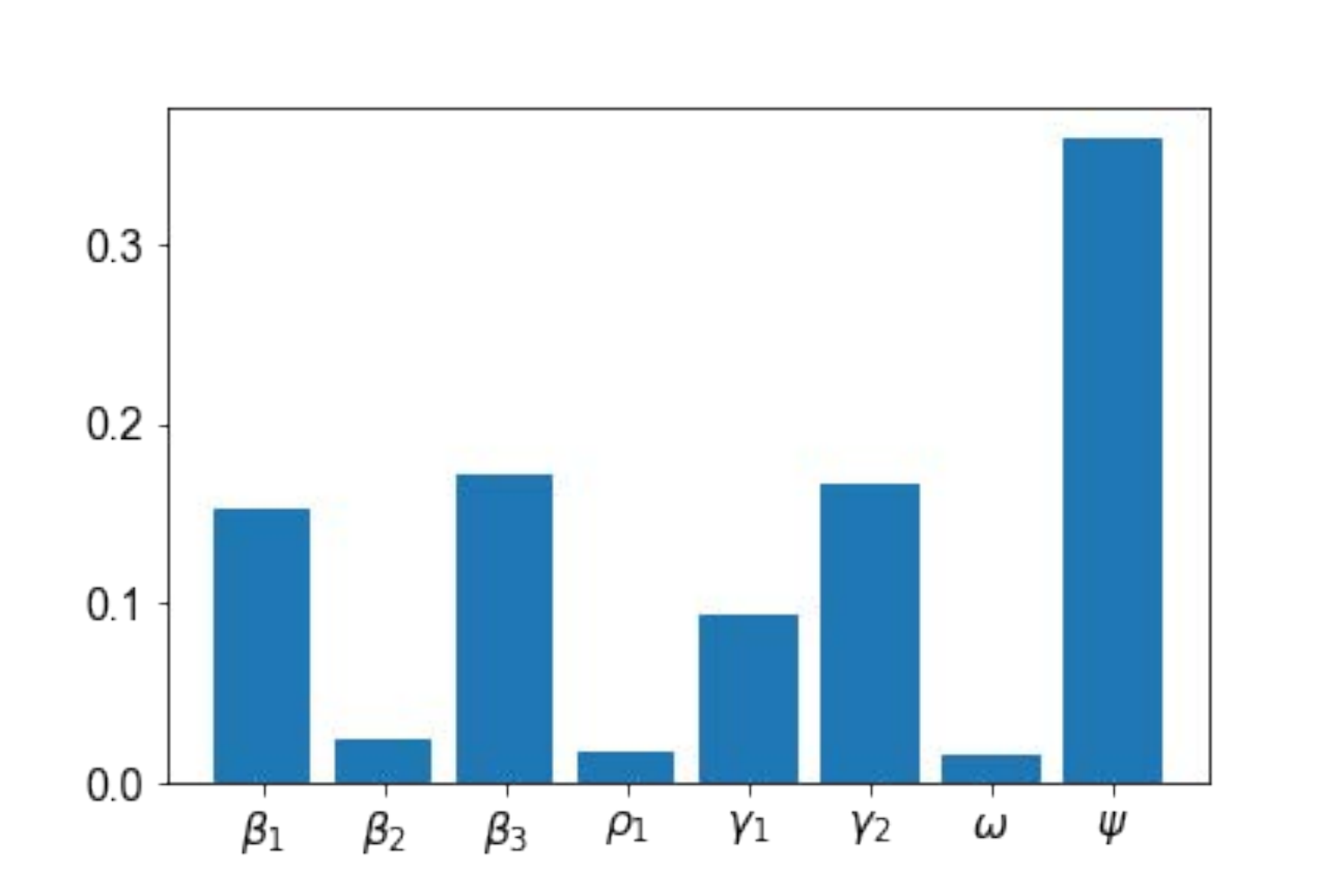}
    \subcaption{DerSHAP values - Liberia}
\end{subfigure}
\begin{subfigure}[b]{.45\textwidth}
    \centering
    \includegraphics[width=1\textwidth]{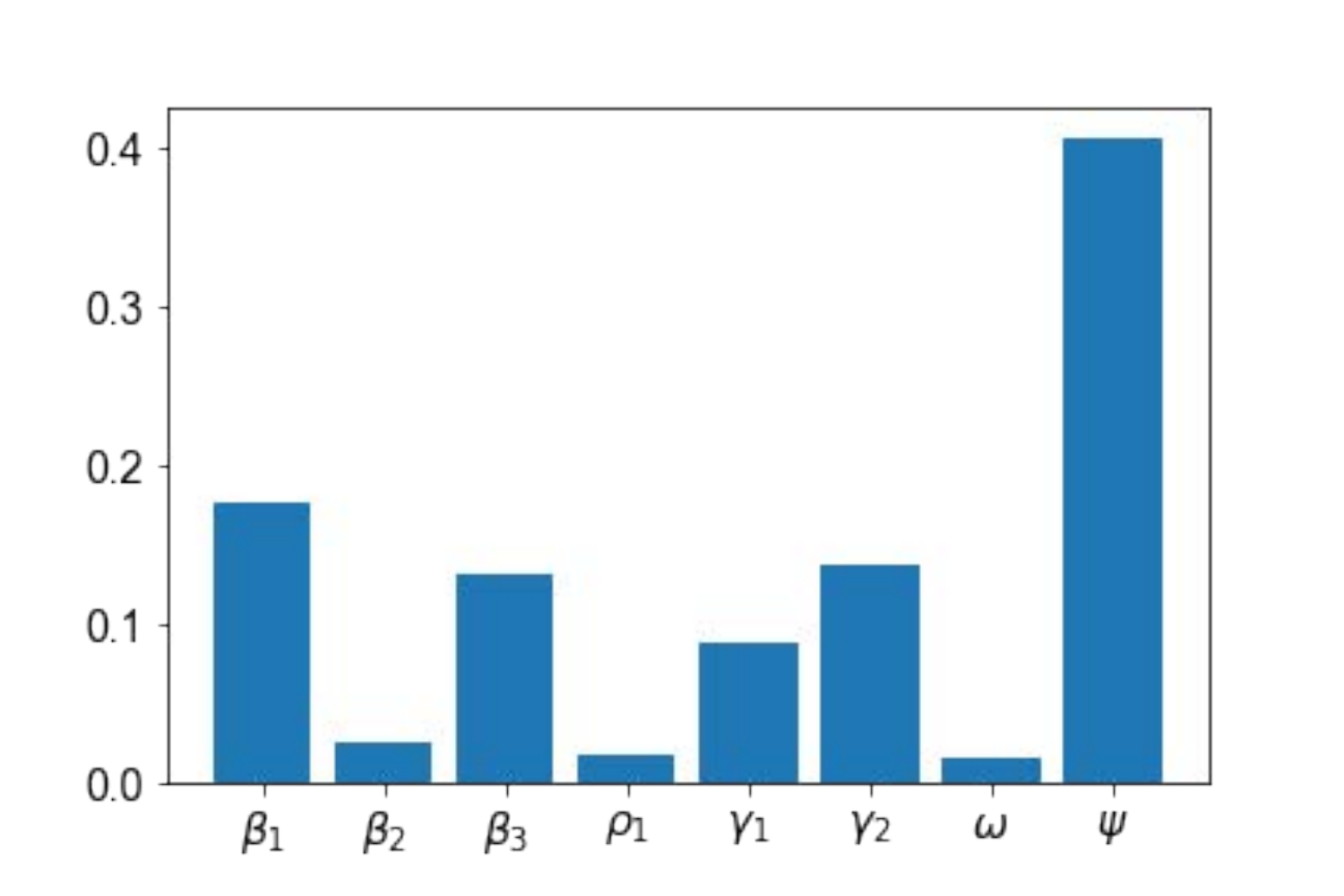}
    \subcaption{DerSHAP values - Sierra Leone}
\end{subfigure}
\caption{Normalized activity scores and DerSHAP values for the Ebola model}
\label{Ebola_plots}
\end{figure}

Table \ref{ebola}, which is from \cite{diaz2018modified}, displays the model parameters and their distributions, obtained using data from Liberia and Sierra Leone. 
\begin{table}[h]
    \centering
    \begin{tabular}{c|c|c}
   Parameter&	Liberia&	Sierra Leone \\
   \hline
$\beta_1$ &	$U(.1, .4)$&	$U(.1, .4)$\\
$\beta_2$ &	$U(.1, .4)$ &	$U(.1, .4)$\\
$\beta_3$ & $U(.05, .2)$ &	$U(.05, .2)$\\
$\rho_1$ &	$U(.41, 1)$ &	$U(.41, 1)$\\
$\gamma_1$&	$U(.0276, .1702)$ &	$U(.0275, .1569)$\\
$\gamma_2$&	$U(.081, .21)$ & $U(.1236, .384)$ \\
$\omega$ &	$U(.25, .5)$ &	$U(.25, .5)$ \\
$\psi$ &	$U(.0833, .7)$ & $U(.0833, .7)$
    \end{tabular}
    \caption{Parameters of Ebola model}
    \label{ebola}
\end{table}
We use the activity scores and DerSHAP to compute the importance of each parameter. The matrix $C$, which is used by both methods, is approximated with Gauss-Legendre quadrature with eight points in each of the eight dimensions of the parameter space as done in \cite{diaz2018modified}. Fig. \ref{Ebola_plots} plots the activity scores and DerSHAP values for the model parameters for the Liberia and Sierra Leone datasets. 

Although the methods agree on the importance of some parameters, there are some significant differences. DerSHAP values for $\beta_2, \rho_1, \omega$ are much larger than their activity scores for both data sets. Parameter $\beta_3$ is less important than $\gamma_2$ according to the activity scores for Sierra Leone data, but they are nearly equally important according to DerSHAP. These differences are due to the fact that DerSHAP takes into account the 2nd order interactions $\partial f_i \partial f_j$ for $i\neq j$, whereas the activity score only considers $\partial f_i ^2$. In terms of computational complexity, the activity score method computes $d$ terms for $d$ inputs, where each term is an expectation, approximated by numerical quadrature (as in this problem) or Monte Carlo, whereas DerSHAP requires the computation of $d$ expectations for one single input and $d+d(d-1)/2$ expectations for all $d$ inputs. 


\subsection{HIV Model}
Here we consider the Human Immunodeficiency Virus (HIV) model in Loudon and Pankavich (2016) \cite{loudon2016}. The HIV model is described by the following differential equations:
\begin{align*}
\frac{dT}{dt} &= s_1 + \frac{p_1}{C_1+V}TV - \delta_1T - (K_1V + K_2M_I)T,\\
\frac{dT_I}{dt} &= \psi(K_1V + K_2M_I)T + \alpha_1T_L-\delta_2T_I-K_3T_ICTL,\\
\frac{dT_L}{dt} &= (1-\psi)(K_1V+K_2M_I)T-\alpha_1T_L-\delta_3T_L,\\
\frac{dM}{dt} &= s_2+K_4MV-K_5MV-\delta_4M,\\
\frac{dM_I}{dt} &= K_5MV-\delta_5M_I-K_6M_ICTL,\\
\frac{dCTL}{dt} &= s_3 + (K_7T_I+K_8M_I)CTL-\delta_6CTL,\\
\frac{dV}{dt} &= K_9T_I+K_{10}M_I-K_{11}TV-(K_{12}+K_{13})MV-\delta_7V,
\end{align*}
where $T(t)$ is the CD4$^+$ T-cell population, $T_I$ is the actively infected T-cell population, $T_L$ represents latently-infected T-cells, $M$ is macrophages, $M_I$ is infected macrophages, $CTL$ is cytotoxic lymphocytes, and $V$ is virions. This model has 27 input parameters 
$$
{\small \textbf{x}=(s_1, s_2, s_3, p_1, c_1, s_1,K_1, K_2, \cdots, K_{13}, \delta_1,\delta_2, \cdots,\delta_7, \alpha_1, \psi).}
$$
The output of the model is the T-cell population, $T(t)$. The distributions of the parameters are given in Loudon and Pankavich (2016) \cite{loudon2016}. The authors considered T-cell count at many time points, and used the active subspace method to obtain a reduced model at each time. Here we will fix one particular time, $t=24$ days, and apply the activity scores and DerSHAP to identify the important input parameters.

\begin{figure}[h]
\centering
\begin{subfigure}[b]{.95\textwidth}
    \centering
    \includegraphics[width=1\textwidth]{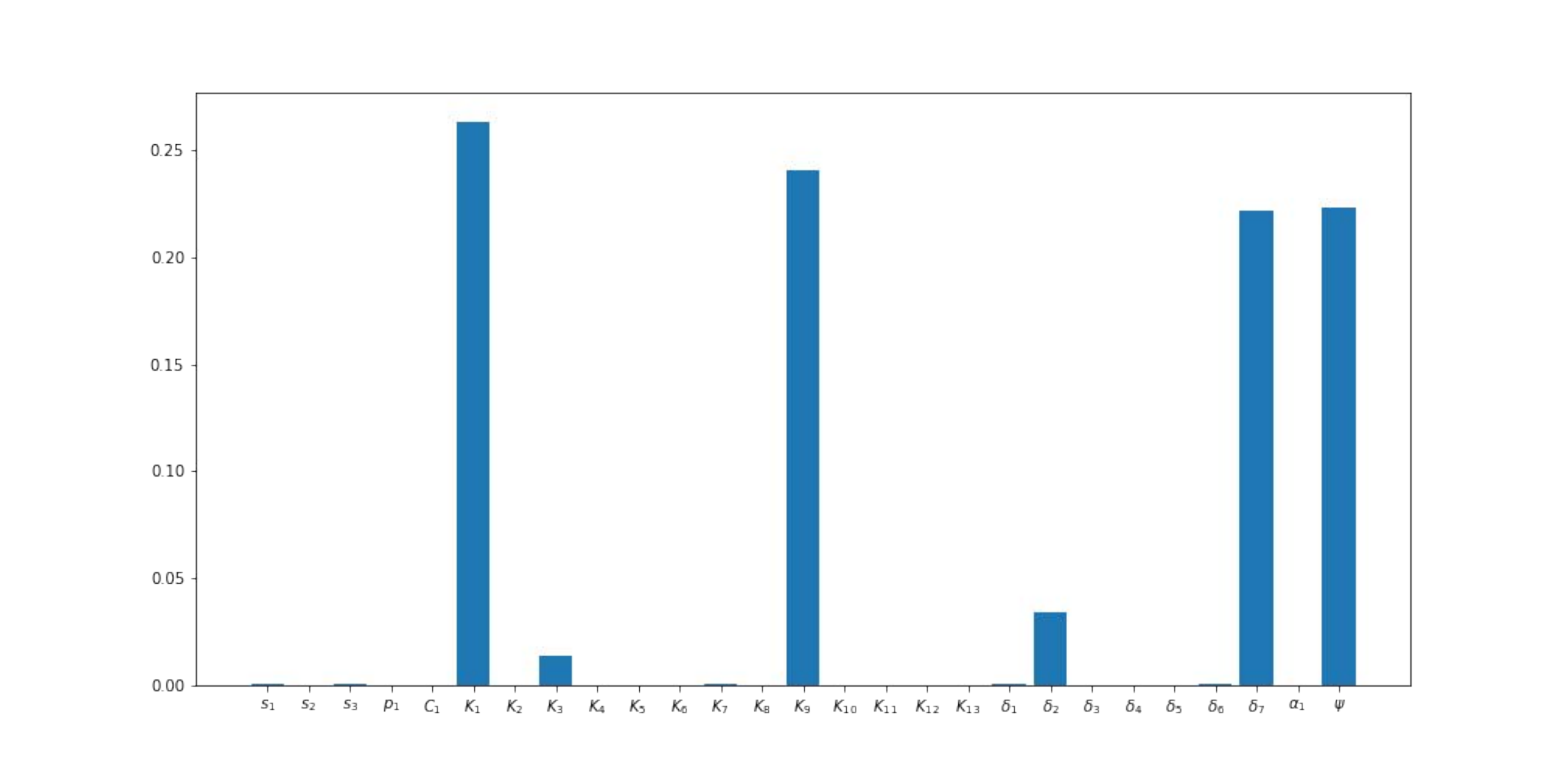}
    \subcaption{Activity scores}
\end{subfigure}%
\\
\begin{subfigure}[b]{.95\textwidth}
    \centering
    \includegraphics[width=1\textwidth]{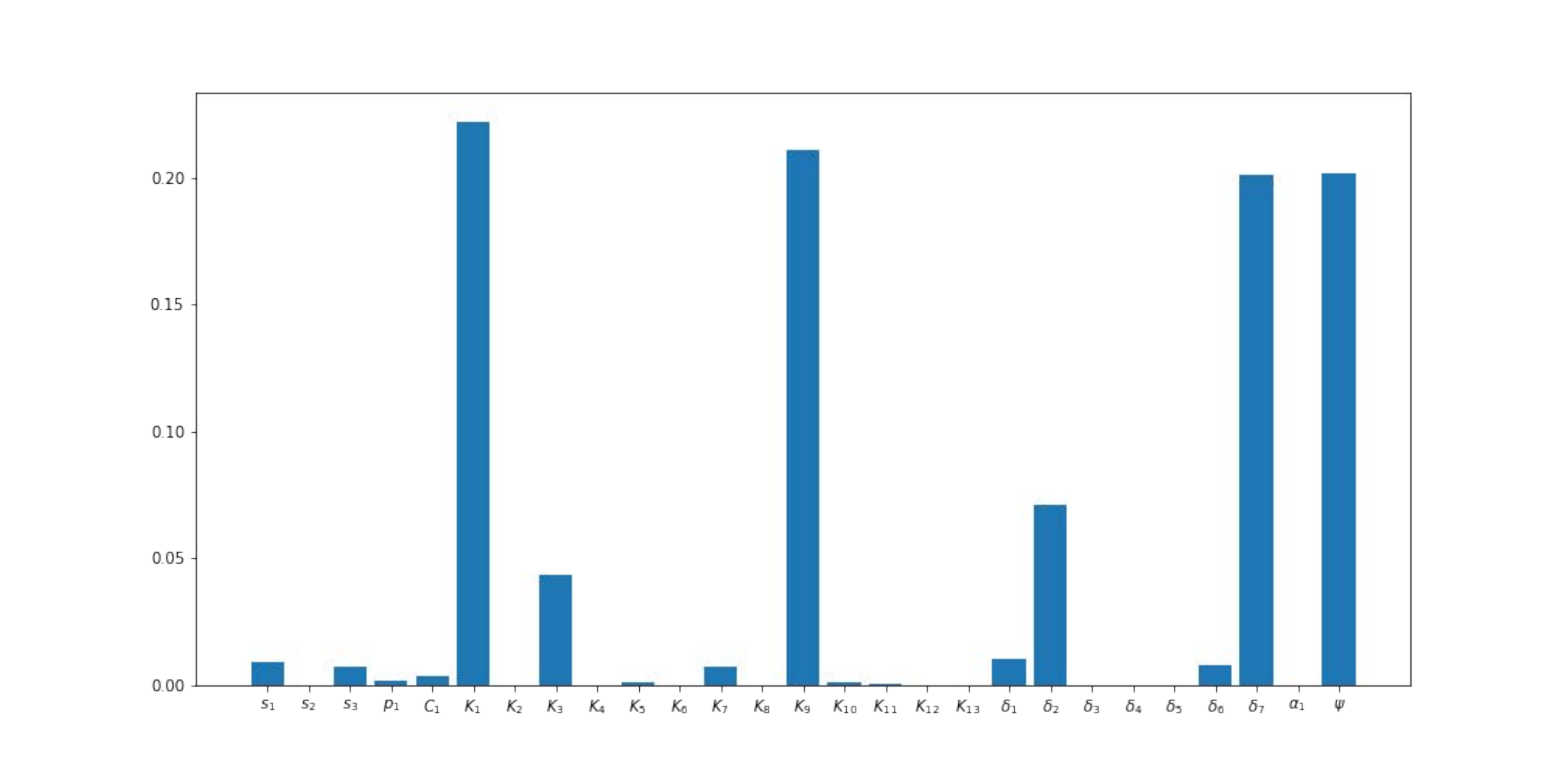}
    \subcaption{DerSHAP values}
\end{subfigure}
\caption{Normalized activity scores and DerSHAP values for the HIV model}
\label{hiv_model}
\end{figure}

Fig. \ref{hiv_model} plots the activity scores and DerSHAP values for the HIV model. Both methods agree on the most important inputs: $(K_1, K_9, \delta_7, \psi)$. However, DerSHAP values for many other input variables are nonzero. In particular, the parameter $\delta_2$ has an activity score of less than $5\%$, but a DerSHAP value of more than $5\%$. The level $5\%$ is important since some researchers suggest freezing input parameters that have importance less than $5\%$. At this threshold, the Shapley value approach would not freeze $\delta_2$, but the activity score approach would.



\subsection{ Amazon Stock Data}
The data, which is from Kaggle\footnote{\url{https://www.kaggle.com/code/gcmadhan/amazon-stock-prediction-nn-fbprophet}}, is the Amazon stock price data from Jan. 4, 2010 to Dec 30, 2016, with 1233 observations in total. The training data contains the opening price, closing price, low price, high price and volume of the stock with 863 observations. The problem is to estimate the closing price (the target), from the remaining data (the features), using a machine learning algorithm. In the notation of the previous sections, the output of the model, $f(\textbf{x})$, is the closing price, and $\textbf{x}$ is the input features. In this example we use a linear regression model to obtain $f(\textbf{x})$, and use the Amazon stock data to train the model. 


Table \ref{amazon_corr} displays the correlation between each input feature and closing price for the whole data. The opening price, low price, and high price have a large correlation with the stock's closing price. The volume of the stock has a smaller negative correlation. 

\begin{table}[h]
    \centering
    \begin{tabular}{c|c}
    Features & Correlation with target\\
    \hline
        opening (Feature 0)  &    0.9996\\
low (Feature 1)    &  0.9998\\
high (Feature 2)   &  0.9998\\
volume (Feature 3)  & -0.2386
    \end{tabular}
    \caption{Correlation between input features and target (closing price)}
    \label{amazon_corr}
\end{table}

\begin{figure}[h]
\centering
\begin{subfigure}[b]{.45\textwidth}
    \centering
    \includegraphics[width=1\textwidth]{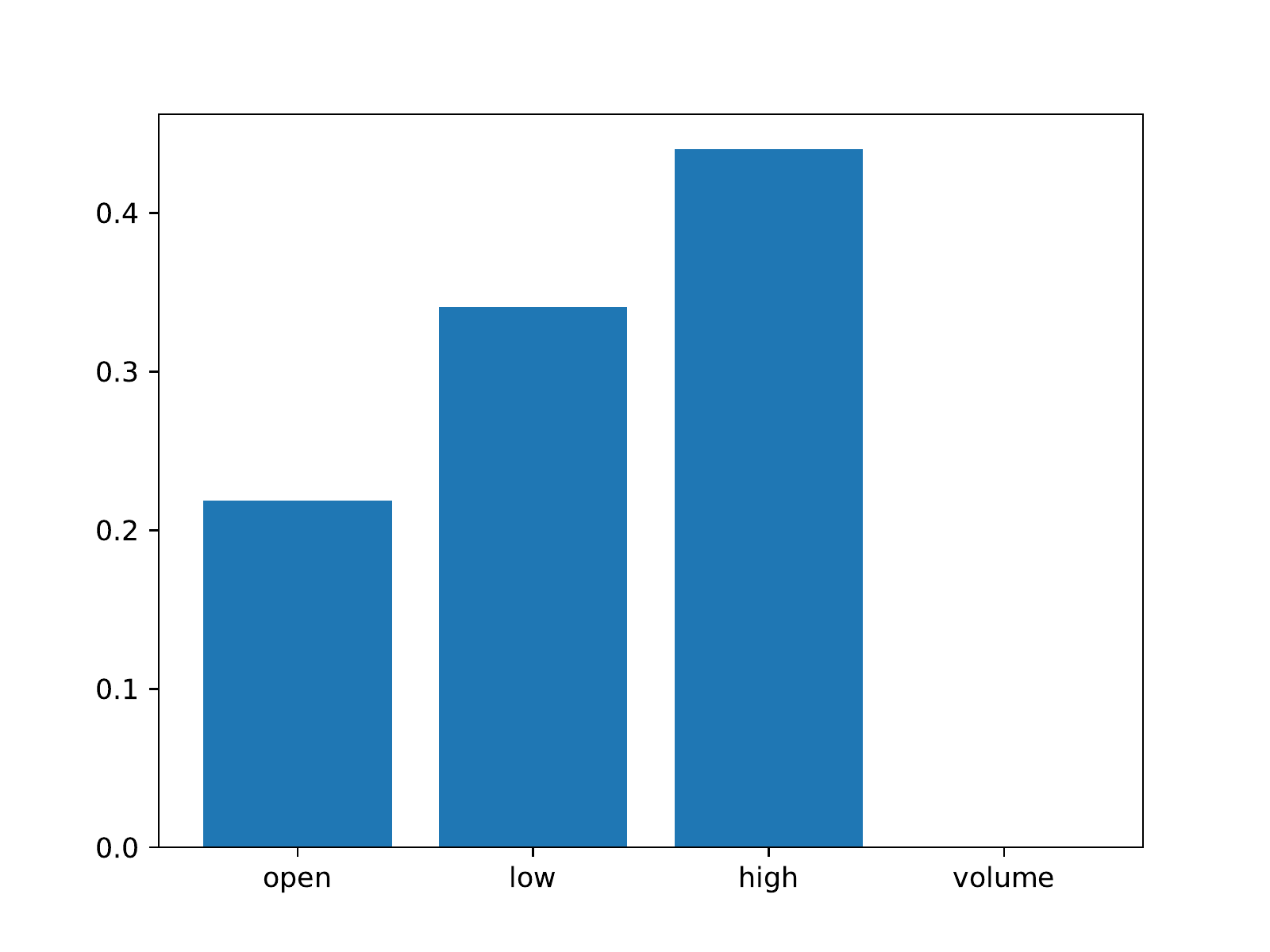}
    \subcaption{DerSHAP}
\end{subfigure}%
\begin{subfigure}[b]{.5\textwidth}
    \centering
    \includegraphics[width=1\textwidth]{figs_pdf/amaz_SHAP4_dep}
    \subcaption{SHAP}
\end{subfigure}
\caption{DerSHAP and SHAP values for dependent Amazon stock data}
\label{amaz_4}
\end{figure}

In the numerical results, we assume the input features are dependent and follow the normal distribution. We use the Python package linear SHAP\footnote{\url{https://shap.readthedocs.io/en/latest/generated/shap.explainers.Linear.html}} by Lundberg, which claims to compute SHAP values accounting for correlations. We compute the expectations $\mathbb{E}[\partial f_i^2]$ and $\mathbb{E}[\partial f_i\partial f_j]$ in DerSHAP using Monte Carlo, based on 1,000 random normal variates where the covariance matrix and the mean vector are obtained from the data. We use forward differences with an increment of $10^{-6}$ to estimate the partial derivatives. Fig. \ref{amaz_4} plots the DerSHAP and SHAP values for the linear regression model. Both methods agree on the importance of inputs: the most important ones are high price, low price, open price, and the least important one is volume.





Table \ref{time_stock} presents the computational time\footnote {Both methods are implemented using Python, on a MacBook Pro M2 Chip. We use the code given by \cite{lundberg2017unified} for SHAP.} to obtain the results in Fig. \ref{amaz_4}. The computational time for both methods are similar. 

\begin{table}[h]
    \centering
    \begin{tabular}{c|c}
    Method  & Time(seconds)\\
    \hline
     DerSHAP    & 0.18 \\
     SHAP & 0.22\\
     \hline
    \end{tabular}
        \caption{Computation time comparison - Amazon stock data}
    \label{time_stock}
\end{table}

\subsection{Boston Housing Data}
The data, which is from Kaggle\footnote{\url{https://www.kaggle.com/c/boston-housing}}, consists of 489 data points where there are eight features: LSTAT (\% lower status of the population), INDUS (proportion of non-retail business acres per town), NOX (nitric oxides concentration (parts per 10 million)), PTRATIO (pupil-teacher ratio by town), RM (average number of rooms per dwelling, TAX (full-value property-tax rate per \$10,000), DIS (weighted distances to five Boston employment centres), AGE (proportion of owner-occupied units built prior to 1940). The target, $f(\textbf{x})$, is MEDV (median value of owner-occupied homes in \$1000s). Using the data we train two different models, support vector regression (SVR) and gradient boosting regression (GBR). The mean and standard deviation of the mean square error for the SVR model is $-0.04 (+/- 0.03)$ and the GBR model is $-0.03 (+/- 0.02)$.

\begin{figure}[h]
\centering
\begin{subfigure}[b]{.45\textwidth}
    \centering
    \includegraphics[width=1\textwidth]{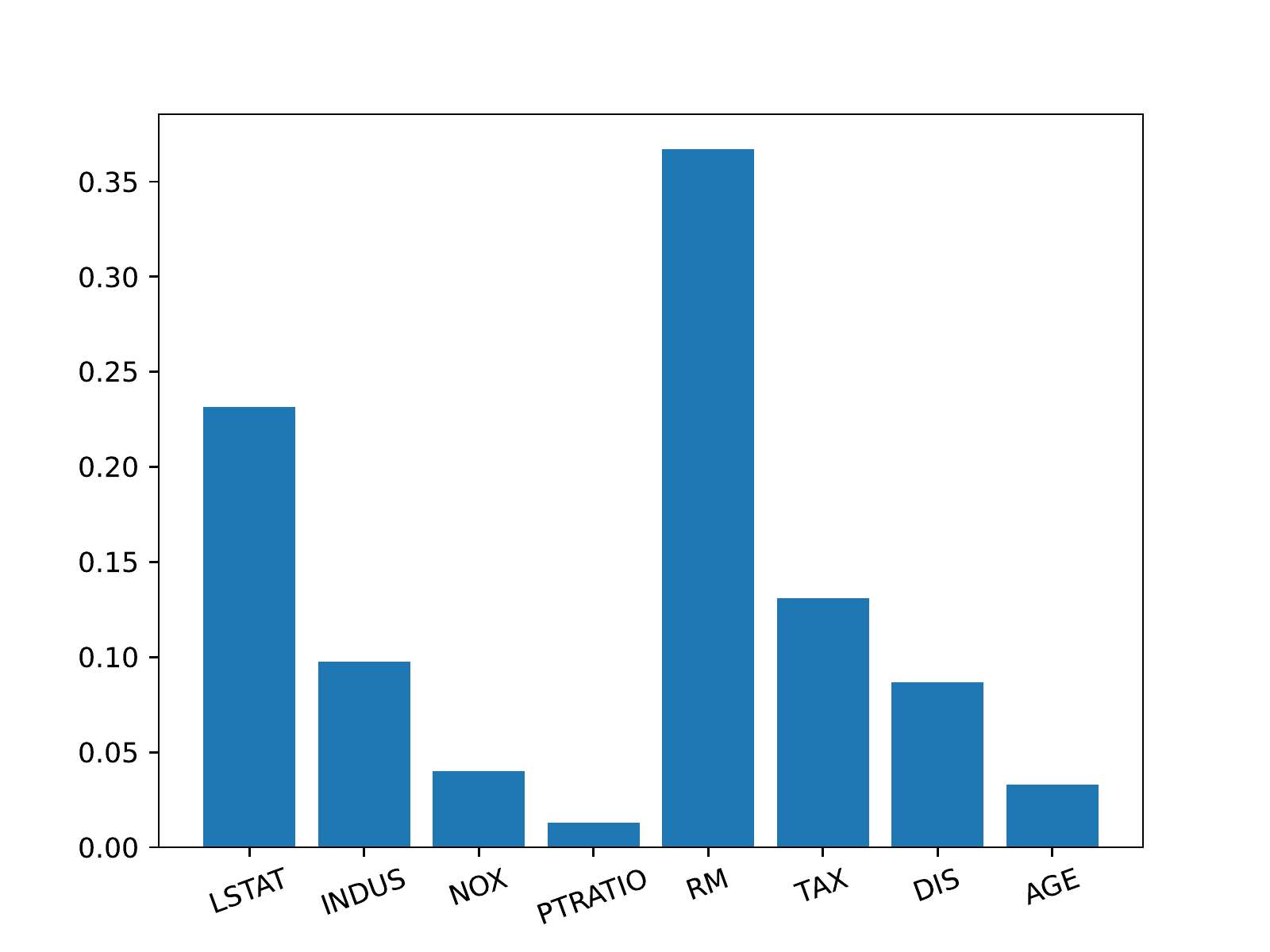}
    \subcaption{DerSHAP}
\end{subfigure}%
\begin{subfigure}[b]{.5\textwidth}
    \centering
    \includegraphics[width=1\textwidth]{figs_pdf/boston_SVM_SHAP_indep.pdf}
    \subcaption{KernelSHAP}
\end{subfigure}
\caption{DerSHAP and SHAP values for Boston housing data - SVR Model}
\label{boston_SVR_indep}
\end{figure}

For DerSHAP, we first assume the input features are independent and follow normal distribution, with parameters estimated from data. We use 10,000 random samples to estimate the expectations, and use forward differences with an increment of $10^{-6}$ to estimate the partial derivatives. For KernelSHAP, we first set the training sample for the explainer to 50. Fig. \ref{boston_SVR_indep} plots the DerSHAP and KernelSHAP values for the SVR model. Both methods agree on the most important inputs: RM, LSTAT, and the three least important inputs (less than 5\% in DerSHAP): NOX, AGE, PTRATIO. The third and fourth important inputs for DerSHAP are TAX and INDUS, whereas for KernelSHAP, they are INDUS and DIS. If we increase the training sample for KernelSHAP from 50 to 490, the third and fourth important variables for KernelSHAP matches the results of DerSHAP. The computing time for KernelSHAP for 490 samples is, however, 3.7 hours, whereas DerSHAP gets the same information in 6.81 seconds. 

\begin{figure}[h]
\centering
\begin{subfigure}[b]{.45\textwidth}
    \centering
    \includegraphics[width=1\textwidth]{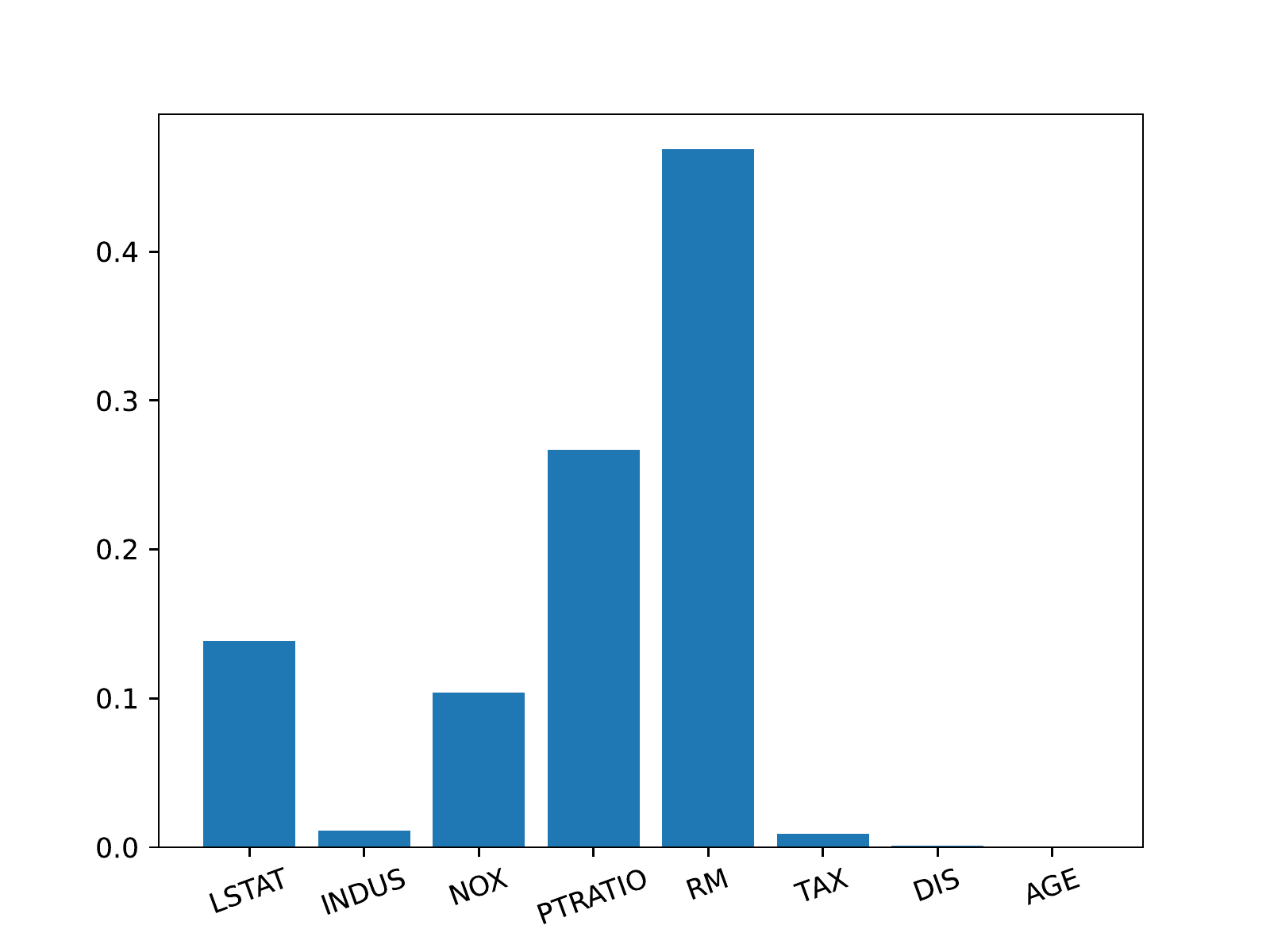}
    \subcaption{DerSHAP}
\end{subfigure}%
\begin{subfigure}[b]{.5\textwidth}
    \centering
    \includegraphics[width=1\textwidth]{figs_pdf/boston_GBR_SHAP_indep.pdf}
    \subcaption{KernelSHAP}
\end{subfigure}
\caption{DerSHAP and SHAP values for Boston housing data - GBR Model}
\label{boston_GBR_indep}
\end{figure}

Fig. \ref{boston_GBR_indep} plots the DerSHAP and KernelSHAP values for the GBR model. DerSHAP identifies the most important inputs as: RM, PTRATIO and LSTAT. However, KernelSHAP identifies the most important inputs as: LSTAT, RM, PTRATIO. DIS has a DerSHAP value of nearly zero, but its KernelSHAP value is not negligible. The conclusions of DerSHAP and KernelSHAP are clearly different, and the differences remain even if we raise the size of the training sample for KenelSHAP to 490 (unlike the SVR case, where KernelSHAP converged to the outcomes of DerSHAP.)

Table \ref{time_housing} presents the computational time of DerSHAP and KernelSHAP with the SVR and GBR models. The computational time for KernelSHAP is about factors of 122 and 539 higher than DerSHAP. Recall that KernelSHAP uses a training sample of size 50, and DerSHAP uses a Monte Carlo sample size of 10,000, for each model, in these comparisons.

\begin{table}[h]
    \centering
    \begin{tabular}{c|c|c}
   Model &Method  & Time(seconds)\\
\hline
  SVR&       DerSHAP  &  6.81\\
    & KernelSHAP & 3676.19\\
     \hline
       GBR&       DerSHAP  &  9.05\\
    & KernelSHAP & 1106.08\\
    \end{tabular}
    \caption{Computation time comparison - Boston housing problem}
    \label{time_housing}
\end{table}

\begin{figure}[h]
\centering
\begin{subfigure}[b]{.49\textwidth}
    \centering
    \includegraphics[width=1\textwidth]{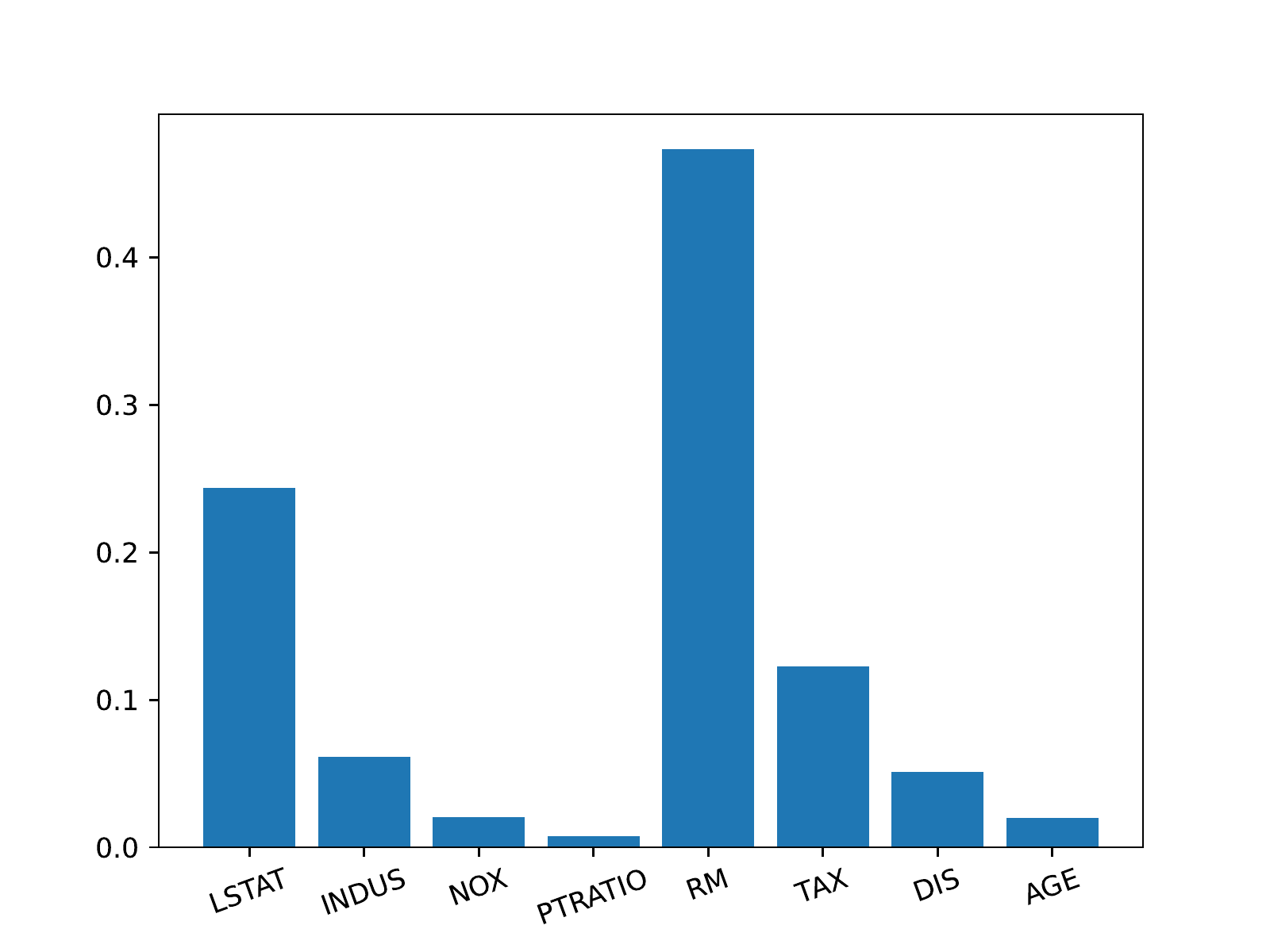}
    \subcaption{DerSHAP (dependent inputs - SVR)}
\end{subfigure}%
\begin{subfigure}[b]{.49\textwidth}
    \centering
    \includegraphics[width=1\textwidth]{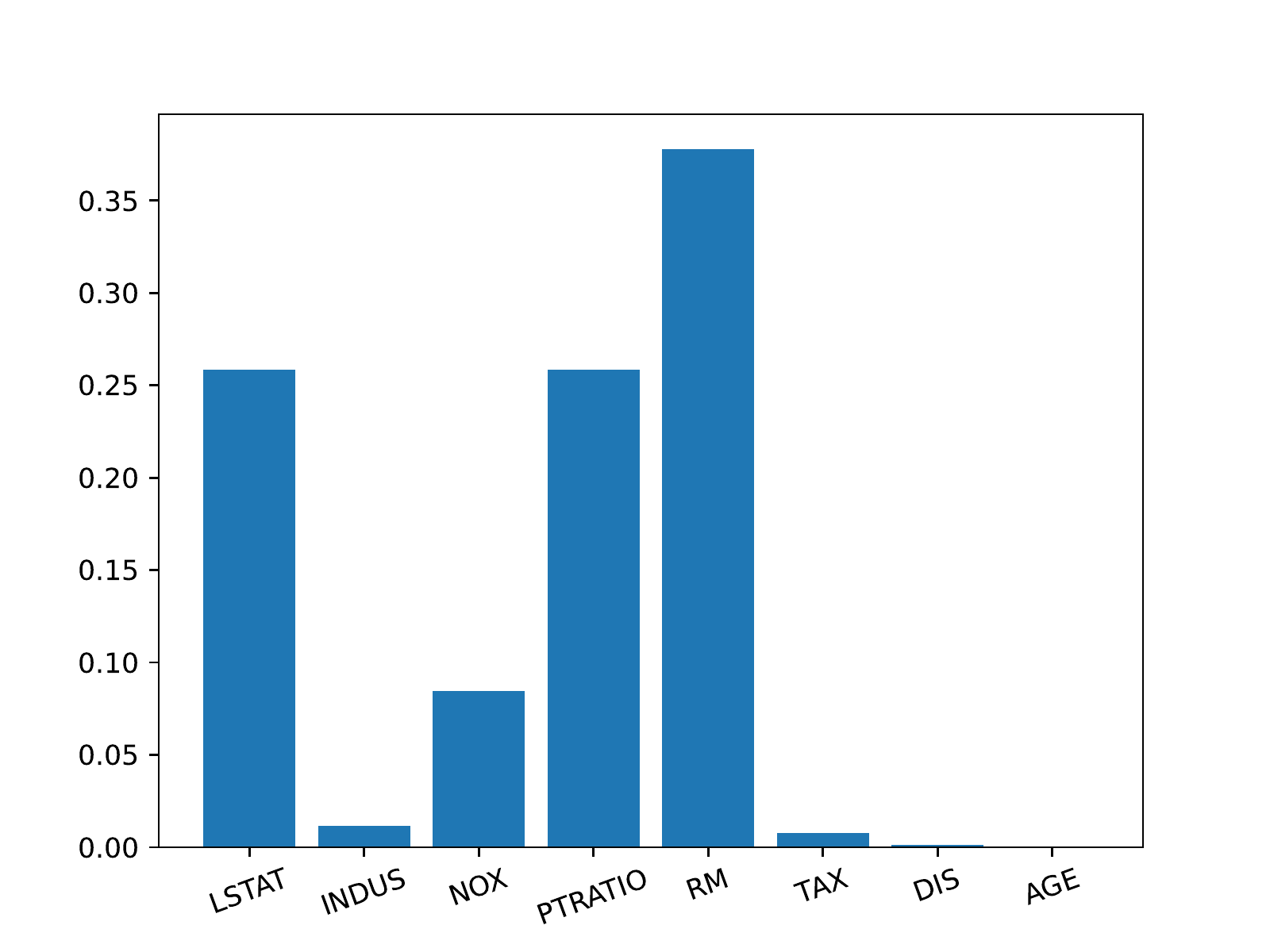}
    \subcaption{DerSHAP (dependent inputs - GBR)}
\end{subfigure}%
\caption{DerSHAP values for dependent inputs for Boston housing data for SVR and GBR models.}
\label{boston_dep}
\end{figure}

An advantage of DerSHAP is it can be used with dependent data, unlike KernelSHAP. Next we model the input features as correlated normal random variables, which is the appropriate choice, using the data to estimate the correlation matrix and the mean vector. Fig. \ref{boston_dep} displays the DerSHAP values for the dependent inputs, for the SVR and GBR models. 
The importance ranking of features under DerSHAP is similar between the independent and dependent input assumptions, with one notable exception. In the case of independence assumption with the SVR model, the DerSHAP value for DIS is 6.9\%, however, when we model features as dependent inputs, the value for DIS becomes 4.3\% - a negligible value for input freezing. 

\section{Conclusions}
DerSHAP is computationally efficient: it has linear complexity as opposed to exponential complexity of other general purpose Shapley algorithms. It can be used with independent and dependent inputs. Unlike DGSM and activity score, DerSHAP incorporates 2nd order interactions via mixed partial derivatives. The efficiency, flexibility, and the ability to include higher order interactions make DerSHAP a competitive method for global sensitivity analysis and machine learning explainability.


\bibliographystyle{siam}
\bibliography{references}
\end{document}

%% file: ex_shared.tex
\usepackage{lipsum}
\usepackage{amsfonts}
\usepackage{graphicx}
\usepackage{epstopdf}
\usepackage{algorithm}
\usepackage{algpseudocode}
\usepackage{subcaption}
\ifpdf
  \DeclareGraphicsExtensions{.eps,.pdf,.png,.jpg}
\else
  \DeclareGraphicsExtensions{.eps}
\fi

\DeclareMathOperator{\x}{\textbf{x}}

\newsiamremark{remark}{Remark}
\newsiamremark{hypothesis}{Hypothesis}
\crefname{hypothesis}{Hypothesis}{Hypotheses}
\newsiamthm{claim}{Claim}

\headers{Derivative-based Shapley value}{Hui Duan and Giray \"{O}kten}

\title{Derivative-based Shapley value for global sensitivity analysis and machine learning explainability}

\author{Hui Duan\thanks{Department of Mathematics, Florida State University, Tallahassee FL, USA
  (\email{hduan@math.fsu.edu}).}
\and Giray \"{O}kten\thanks{Department of Mathematics, Florida State University, Tallahassee FL, USA
  (\email{okten@fsu.edu}).}
}

\usepackage{amsopn}
\DeclareMathOperator{\diag}{diag}

\makeatletter
\newcommand*{\addFileDependency}[1]{
  \typeout{(#1)}
  \@addtofilelist{#1}
  \IfFileExists{#1}{}{\typeout{No file #1.}}
}
\makeatother

\newcommand*{\myexternaldocument}[1]{%
    \externaldocument{#1}%
    \addFileDependency{#1.tex}%
    \addFileDependency{#1.aux}%
}